\documentclass[runningheads]{llncs}
\usepackage[T1]{fontenc}
\usepackage{graphicx}
\usepackage{multirow}
\usepackage{array}
\usepackage{booktabs}
\usepackage{multirow}
\usepackage{xcolor}

\usepackage[colorlinks=true,  
            linkcolor=blue,   
            citecolor=blue,   
            urlcolor=blue]{hyperref}
\usepackage{amsmath}
\usepackage{wrapfig}
\usepackage{lipsum}
\usepackage{amsfonts}
\usepackage{amssymb}
\usepackage{algorithm}
\usepackage{pifont}
\usepackage[noend]{algpseudocode}
\usepackage{graphicx}

\newcommand{\modelacro}{{NeuroPathX }}
\newcommand{\modelacroNSP}{{NeuroPathX}}

\newcommand{\wo}{w\!/o}
\usepackage{xcolor}
\begin{document}
\title{Learning Explainable Imaging-Genetics Associations Related to a Neurological Disorder}
%
%

%
\author{Jueqi Wang\inst{1} \and
Zachary Jacokes\inst{2} \and
John Darrell Van Horn\inst{2,3} \and
Michael C. Schatz\inst{4} \and
Kevin A. Pelphrey\inst{5} \and
Archana	Venkataraman\inst{1}
}
\authorrunning{J. Wang et al.}
\titlerunning{Explainable Imaging-Genetics Associations}
%
\institute{Department of Electrical and Computer Engineering, Boston University \\
\email{\{jueqiw,archanav\}@bu.edu} \and
 School of Data Science, University of Virginia \\
\email{zj6nw@virginia.edu} \and
Department of Psychology, University of Virginia \\
\email{jdv7g@virginia.edu} \and
 Departments of Computer Science and Biology, Johns Hopkins University \\
\email{mschatz@cs.jhu.edu} \and
 Department of Neurology, University of Virginia \\
\email{kevin.pelphrey@virginia.edu}
 }
\maketitle              
\begin{abstract}
While imaging-genetics holds great promise for unraveling the complex interplay between brain structure and genetic variation in neurological disorders, traditional methods are limited to simplistic linear models or to black-box techniques that lack interpretability. In this paper, we present \modelacroNSP, an explainable deep learning framework that uses an early fusion strategy powered by cross-attention mechanisms to capture meaningful interactions between structural variations in the brain derived from MRI and established biological pathways derived from genetics data. To enhance interpretability and robustness, we introduce two loss functions over the attention matrix -- a sparsity loss that focuses on the most salient interactions and a pathway similarity loss that enforces consistent representations across the cohort. We validate \modelacro on both autism spectrum disorder and Alzheimer's disease.
Our results demonstrate that \modelacro outperforms competing baseline approaches and reveals biologically plausible associations linked to the disorder. 
These findings underscore the potential of \modelacro to advance our understanding of complex brain disorders. 

\keywords{Imaging-genetics  \and Neurological disorder \and Multimodal fusion \and Interpretability \and Biologically-inspired deep learning}
\end{abstract}

\section{Introduction}

Neurological disorders are marked by considerable heterogenity, which is in part driven by inherited genetic factors~\cite{jack2021neurogenetic,buch2023molecular}. These genetic influences can not only modulate disease susceptibility but can also shape the trajectory of disease progression. In parallel, brain imaging offers a unique window into how genetic variations may affect brain structure and function \cite{pretzsch2023structural,wen2016pathway}. Thus, integrating neuroimaging and genetics data promises a more comprehensive framework to unravel the complex interplay between neural and biological pathways, which in turn, will advance our understanding of multifaceted and heritable disorders.

A promising avenue for exploring the imaging-genetic interplay is brain imaging transcriptomics, which measures gene expression in brain tissue samples~\cite{hawrylycz2012anatomically,gtex2020gtex}. However, these studies rely on postmortem data from a small number of donors. Thus, they cannot capture dynamic processes in the brain both in time and across neurodevelopment~\cite{arnatkevic̆iute2019practical}. In contrast, traditional \textit{in vivo} imaging-genetics studies have largely modeled linear associations between single-nucleotide polymorphisms (SNPs) and neuroimaging phenotypes \cite{shen2010whole,xu2017imaging}. This simple framework may overlook the complex and nonlinear interactions that are critical to fully understanding the biological mechanisms underlying the disorder~\cite{choi2023prset,smedley2020discovering}.

Machine learning (ML) has emerged as a powerful tool for identifying biomarkers in neurological diseases. Most often, a classifier is trained to distinguish between patients and neurotypical controls, and a post-hoc evaluation of the model is used to reveal novel biomarkers~\cite{chen2021multimodal,dvornek2023copy,jaume2024modeling,smedley2020discovering,zhou2023integrating,ghosal2021g}. While ML models excel in capturing nonlinear interactions, they are often applied to a small subset of SNP values~\cite{zhou2023integrating,ghosal2021g,10313442}. 
This approach is limiting, as the heritability of neurological traits is thought to result from the additive effects of hundreds of SNPs spread throughout the genome and genes functioning as part of intricate pathways~\cite{pretzsch2023structural}.

A few works have explored deep learning for imaging-genetics by a \textit{latent fusion} method, i.e., by combing the low-dimensional encoding for each modality~\cite{zhou2023integrating,ghosal2021g,10313442}. The interpretability of these models are derived from a learned feature importance score. However, the latent fusion approach may not capture direct interactions between specific brain regions and SNPs. More recent studies have explored the the \textit{early-fusion} method, which models cross-modal interactions between the input features themselves~\cite{jaume2024modeling,chen2021multimodal,smedley2020discovering}. This is done by using the cross-attention layer inside a transformer architecture to learn the interaction between the imaging and genetics data. However, these models are unable to control or to enforce consistency in the learned interactions across subjects. In addition, the proposed transformer architecture~\cite{jaume2024modeling} uses a genetic encoder that contains a large number of parameters. Finally, the work of~\cite{ghosal2022biologically} takes the unique approach of constructing a graph neural network based on an established ontology of biological pathways~\cite{ashburner2000gene}. However, this model operates at the gene level, which requires a large training sample and is prone to overfitting.


We propose \modelacroNSP, an explainable AI framework that uses cross-atten-tion to capture the intricate interplay between structural brain variation and genetic influences. To better align with the complexity of neurological disorders, \modelacro aggregates SNPs into well-established biological pathways \cite{kanehisa2000kegg}, with each representing a network of genes that describes a specific cellular and/or metabolic function. Hence, \modelacro can deliver interpretable insights into the underlying neurobiology of a particular disorder and can learn effectively on small multimodal datasets. We promote interpretability
via a pairwise loss term that enforces pathway-level coherence across subjects. We demonstrate \modelacro on two distinct datasets, one of autism spectrum disorder (ASD) and another of Alzheimer's disease (AD). Our experiments underscore the robustness, interpretability, and predictive power of our proposed framework.

\begin{figure}[t]\includegraphics[width=0.95\textwidth]{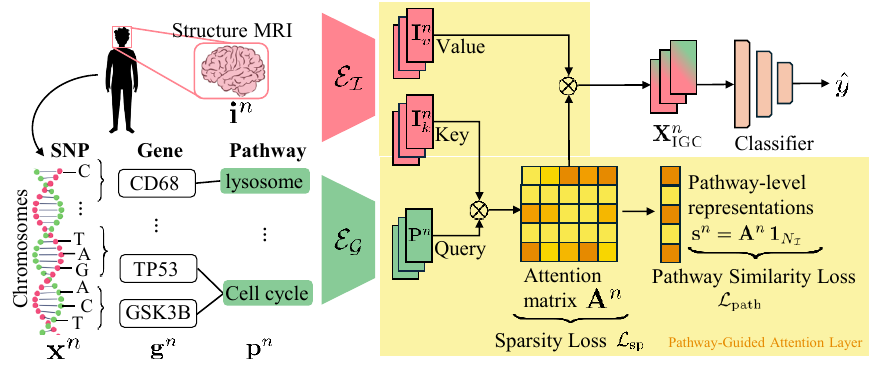}
    \caption{\modelacro overview. Our framework integrates genetic variations (SNPs) and brain MRI features using a pathway-guided attention layer (yellow panel) to capture interactions. We use a sparsity and a pathway similarity loss to enhance interpretability.}
    \label{fig:ModelOverview}
\end{figure}


\section{Pathway-driven Multimodal Fusion and Interpretability}

Fig.~\ref{fig:ModelOverview} illustrates our proposed \modelacro framework. The inputs are a collection of imaging features extracted from brain MRI and a set of genetic features that that encapsulate different biological processes. By employing a cross-attention layer for early fusion, the model not only effectively combines the two modalities but also offers interpretability through the analysis of the attention matrix. 
Formally, let~$\mathbf{i}^n \in \mathbb{R}^{d \times N_{\mathcal{I}}}$ be the input imaging feature vector for subject~$n$, where $N_{\mathcal{I}}$ is the total number of ROIs, and $d$ is the number of features extracted per ROI. The input construction from genetics data is described below. 

\subsection{Biological Pathway Features from SNPs}
We first preprocess the raw SNP data using the conventional genome-wide polygenic risk score (PRS) pipeline~\cite{choi2020tutorial} and select disease-specific SNPs based on a corresponding genome-wide association study (GWAS) from the literature. Let $\mathbf{x}^n \in \mathbb{R}^{N_G}$ denote the SNP vector for subject $n$, where $N_G$ is the number of SNP positions from the GWAS. The selected SNPs are then mapped to genes based on established biological pathway databases~\cite{kanehisa2000kegg}. Specifically, for each gene $i$, we compute a weighted sum of the corresponding SNP values using effect sizes $\beta_j$ (where $j$ is the SNP position from the disease-specific GWAS~\cite{shen2010whole,wightman2021genome}). This process yields \(\mathbf{g}_i^n = \sum_{j \in \Omega_i} \beta_j \, \mathbf{x}_j^n,\) where $\Omega_i$ is the set of SNP indices mapped to gene $i$. We generate a pathway-level representation by aggregating these gene-level features for each pathway $\mathbf{p}_k^n$ as follows: \(\mathbf{p}_k^n = \sum_{l \in \Gamma_k} g_l,\) where $\Gamma_k$ defines the set of genes associated with~pathway $k$~\cite{kanehisa2000kegg}. The pathway vector $\mathbf{p}^n \in \mathbb{R}^{1 \times N_p}$ is the input to 
NeuroPathX, where $N_p$ denotes the total number of pathways.

\subsection{Pathway-Guided Attention Layer}
We use a series of (small) modality-specific encoding layers at the input. This transform the genetics input~$\mathbf{p}^n$ into a representation~$\mathbf{P}^n \in \mathbb{R}^{N_{\mathcal{P}} \times d_{q}}$, where $d_q$ is the pathway encoding dimension. For the imaging data, we construct~$\mathbf{I}^n_k \in \mathbb{R}^{N_{\mathcal{I}} \times d_{k}}$ and~$\mathbf{I}^n_v \in \mathbb{R}^{N_{\mathcal{I}} \times d_v}$ with encoding dimensions~$d_k$ and~$d_v$, respectively. 
Unlike a traditional cross-attention mechanism, in which \(\mathbf{I}_v^n\) and \(\mathbf{I}_k^n\) are identical, we design separate encoding layers to learn more flexible associations. In addition, we preserve the input feature dimensionality to learn the direct interactions between individual genetic pathways and brain ROIs. The encoded representations are then projected via three learnable linear mappings \(\mathbf{W}_q\in \mathbb{R}^{d_q \times d_{qk}}\), \(\mathbf{W}_k\in \mathbb{R}^{d_{k} \times d_{qk}}\), and \(\mathbf{W}_v\in \mathbb{R}^{d_v \times d}\) for a subsequent cross-attention layer.

Motivated by the works of~\cite{chen2021multimodal,jaume2024modeling}, we derive a Pathway-guided Cross Attention (PathAttn) layer to learn the interactions between imaging and genetics. This layer uses the pathway guidance signal \(\mathbf{P}^n\) to direct the aggregation of imaging features into a clustered set of Imaging-Guided pathway Concepts (IGC), denoted as \(\mathbf{X}^n_{\text{IGC}} \in \mathbb{R}^{N_{\mathcal{I}} \times d}\). Mathematically, this operation is defined as
\begin{align}
    \nonumber
    \text{PathAttn}(\mathbf{I}^n_k, \, \mathbf{I}^n_v, \, \mathbf{P}^n) &= \text{SoftSign}\Big(\text{ReLU}\Big(\frac{\mathbf{P}^n\,\mathbf{W}_q\,\mathbf{W}_k^\top\,(\mathbf{I}^n_k)^\top}{C}\Big)\Big)\mathbf{W}_v\mathbf{I}^n_v \\[2ex] \label{eq:crossattn}
    &= \mathbf{A}^n\mathbf{W}_v\mathbf{I}^n_v \triangleq
    \mathbf{X}^n_{\text{IGC}}
\end{align}
where~$\mathbf{A}^n \in \mathbb{R}^{N_{\mathcal{I}} \times N_{\mathcal{G}}}$ is the pathway-guided attention matrix for subject~$n$ and $\text{SoftSign}(x) = \frac{x}{0.5 + x}$. We use a ReLU activation instead of the traditional softmax function in the cross-attention layer to accommodate the possibility that certain pathways may not be relevant to the disease and should have uniformly lower attention weights. To stabilize the ReLU output, we use a scaling constant $C = \gamma \sqrt{N_\mathcal{G} / 2}$ introduced by~\cite{shen2023study}. Finally, the imaging-guided pathway concept $\mathbf{X}_{\text{IGC}}$ is forwarded to a feed-forward classifier for disease prediction.

\subsection{Enhancing Interpretability via the Loss Functions}  
We design two complementary loss functions on the attention matrix~$\mathbf{A}^n$ to improve the explainability of \modelacroNSP. These are described below.

\medskip \noindent
\textbf{Sparsity Loss:} Our first loss encourages the attention matrix to focus on the most salient interactions between genetic pathways and brain regions. This is done via a sparsity penalty on the entries~$\mathbf{A}^n_{ij}$ of the attention matrix
\begin{equation} \label{eq:sparse}
    \mathcal{L}_{\text{sp}} = \sum_{i} \sum_{j} \mathrm{KL}\Bigl(\mathrm{Ber}(q) \,\|\, \mathrm{Ber}(\mathbf{A}_{ij})\Bigr)  
\end{equation}
where $\mathrm{Ber}(q)$ is a Bernoulli distribution with~$q$ determining the sparsity level~\cite{ghosal2021g}. Since Eq.~\eqref{eq:sparse} is computationally intensive, we approximate~$\mathcal{L}_{\text{sp}}$ by by applying the loss to two randomly selected subjects from each batch during training.

\medskip \noindent
\textbf{Pathway Similarity Loss:} We hypothesize that there is a shared set of biological pathways associated with a particular disorder, and that these pathways have different influences across the brain ROIs between patients (PAT) and neurotypical controls (NC). We integrate this assumption into \modelacro by designing a novel paired similarity loss that encourages the learned pathway-level representations to be consistent between PAT and NC subjects.

Mathematically, we first sum the attention matrix~$\mathbf{A}^n$ across its columns, i.e., over brain regions, yielding the vector $\mathbf{s}^n = \mathbf{A}^n \, \mathbf{1}_{N_{\mathcal{I}}}$, where $\mathbf{1}_{N_{\mathcal{I}}}$ is the all-ones vector of dimension~$N_{\mathcal{I}}$ and $\mathbf{s}^n \in \mathbb{R}^{1 \times N_{\mathcal{G}}}$ for each subject~$n$. During each training epoch, we randomly sample~$M$ pairs of PAT and NC subjects. For each sampled pair~$m$, we construct the vectors~$\mathbf{s}^{(m)}_{\text{PAT}}$ and~$\mathbf{s}^{(m)}_{\text{NC}}$ from the corresponding attention matrices and define the similarity loss for the batch as:
\begin{equation} \label{eq:path}
    \mathcal{L}_{\text{path}} = \frac{1}{M} \sum_{m=1}^{M} \left\lVert \mathbf{s}_{\text{NC}}^{(m)} - \mathbf{s}_{\text{PAT}}^{(m)} \right\rVert_1,
\end{equation}
where the \(\ell_1\) norm both promotes sparsity and encourages consistent pathway-level representations across the patient and control cohorts. 

Finally, we implement a weighted cross-entropy loss on the end classification network. The overall loss for our model is the sum of these terms:
\begin{equation} \label{eq:final}
    \mathcal{L} = \lambda_{\text{sp}}\,\mathcal{L}_{\text{sp}} + \lambda_{\text{path}}\,\mathcal{L}_{\text{path}} + \sum_{n=1}^N \Bigl[\delta \cdot y^n \log(\hat{y}^n) + (1 - \delta)(1-y^n) \log(1-\hat{y}^n)\Bigr]
\end{equation}
where~$y^n$ and $\hat{y}^n$ are the true and predicted classes of subject~$n$, and $\delta$ controls the class weighting in the cross entropy function. The hyperparameter $\delta$ and the hyperparameters~$\{\lambda_{\text{sp}},\lambda_{\text{path}}\}$ are fixed based on the training data. 

The multifaceted loss function \modelacro not only drives sparse and interpretable attention matrices, but it also ensures consistent pathway selection across the cohort while maintaining robust predictive performance.

\subsection{Implementation Details}
We maintain consistent hyperparameters across both datasets in Section~3. We set the learning rate to $5 \times 10^{-5}$ and the Bernoulli probability to $q = 1 \times 10^{-2}$. The hyperparameters in Eq.~\eqref{eq:final} were fixed at $\lambda_{\text{path}} = 1 \times 10^{-3}$ and $\lambda_{\text{sp}} = 1 \times 10^{-6}$. In each epoch, we sample $M = 10$ (PAT/NC) pairs and employed a batch size of 128. The dimensions for the query, key, and value representations were set to $d_q = 32$, $d_k = 4$, $d_v = 8$, and the projected dimension $d_{qk} = 32$ for both datasets.  \modelacro was trained for 1300 epochs on ACE and 100 epochs on ADNI.

\modelacro\footnote{Code is available at \url{https://github.com/jueqiw/NeuroPathX}} was optimized using AdamW with a weight decay of $1 \times 10^{-5}$ and momentum parameters $\beta_1 = 0.9, \beta_2 = 0.999$ \cite{loshchilov2017decoupled}. Batch normalization was applied during training \cite{ioffe2015batch}. We use an MLP with dropout as the final classifier.

\medskip \noindent
\textbf{Baseline Comparisons}: We evaluate \modelacro against two ablated variants of our model: (1)~omitting the sparsity loss $\mathcal{L}_{\text{sp}}$ and (2)~omitting the pathway similarity loss $\mathcal{L}_{\text{path}}$, and to state-of-the-art methods in the literature. These baselines include G-MIND~\cite{ghosal2021g}, which uses a decoder for regularization and learns a shared latent space between imaging and genetic data, and SurvPATH~\cite{jaume2024modeling}, which captures intermodal and intramodal interactions through a unified cross attention matrix. To mitigate overfitting, we did not implement the pathway embedding in SurvPATH. For consistency, we used the same pathway inputs and imaging feature representations as in \modelacroNSP. Finally, we compared \modelacro with Canonical Correlation Analysis (CCA) as a classical baseline.  

\section{Data and Preprocessing}
We evaluate \modelacro on a publicly available dataset of autism spectrum disorder from an Autism Center of Excellence (ACE)\footnote{Data available from \url{https://nda.nih.gov/edit_collection.html?id=2021}} and on the public Alzheimer's Disease Neuroimaging Initiative (ADNI) dataset~\cite{weiner2017recent}. We remove related samples from both datasets to prevent the inflation of brain-pathway associations and to limit data leakage. This results in N=165 ($98$~ASD, $67$~NC) subjects for ACE and N=438 subjects for ADNI ($168$~AD, $270$~NC).

\medskip \noindent
\textbf{Genetics:} Genotyping for ACE was done using the HumanOmni 2.5M BeadChip and for ADNI was done using the Human610-Quad BeadChip~\cite{saykin2015genetic}. For both datasets, we include only autosomal SNPs with a genotyping rate above 99\%, a Hardy–Weinberg equilibrium p-value larger than \(1\times10^{-6}\), a minor allele frequency above 0.01, and a hard call threshold of at most 0.1.

We compute linkage disequilibrium (LD) matrices from the European population of the 1000 Genomes Project. For LD clumping, we set an $R^2$ threshold of 0.1~\cite{weiner2017polygenic} and a physical distance threshold of 50 kilo-base pairs (KB) for both the ACE and ADNI datasets. After this process, 44,102 SNPs in ACE and 45,074 SNPs in ADNI remained available for SNP-to-gene mapping.

For SNP-to-gene mapping, we assign SNPs located within 50~KB to each gene. These genes are further aggregated according to the KEGG canonical pathway gene sets~\cite{kanehisa2000kegg} to generate the input feature vector~$\mathbf{p}^n$ (see Section~2.1). To better focus on neurological relevance, we omit KEGG pathways associated with cancer. This filtering resulted in $N_{\mathcal{P}} = 177$ pathways, encompassing 3,907 genes for ACE and 3,943 genes for ADNI, each with at least one mapped SNP.

\medskip \noindent
\textbf{Neuroimaging:} We use T1-weighted MRI scans to extract anatomical features from the human brain, with preprocessing performed using FreeSurfer~\cite{fischl2012freesurfer}. 
We parcellate the brain according the the Brainnetome atlas~\cite{li2023brainnetome} ($N_{\mathcal{I}} = 210$ ROIs) for ACE and the Desikan-Killiany atlas~\cite{desikan2006automated} ($N_{\mathcal{I}} = 68$ ROIs) for ADNI. For each ROI, we extract $d=4$ structural features: volume size, surface area, average cortical thickness, and standard deviation of the cortical thickness. The ADNI imaging features were available precomputed as part of the TADPOLE challenge~\cite{marinescu2019tadpole}. All training, validation, and test imaging data were normalized using the mean and standard deviation computed from the training set. 

\section{Experimental Results}
\begin{table}[t]
    \setlength{\tabcolsep}{5pt}
    \centering
    \caption{Quantitative evaluation of disease status prediction by \modelacro and the baseline and ablation on the ACE and ADNI datasets. The best performance for each metric is highlighted in bold. * denotes statistically worse performance than best model. (p < 0.05, paired two-tailed t-test)}
    \label{tab:quantitative_results}    
    {\begin{tabular}{p{0.1cm}l|cccc}
        \toprule
        & Methods & Accuracy $\uparrow$ & Sensitivity $\uparrow$ & Specificity $\uparrow$ & AUC $\uparrow$ \\
        \midrule
        \multirow{6}{*}{\rotatebox[origin=c]{90}{ACE}} 
        & CCA & 0.557$\pm$0.07* & \textbf{0.900$\pm$0.10} & 0.060$\pm$0.10* & 0.478$\pm$0.06* \\
        & G-MIND & 0.569$\pm$0.09* & 0.593$\pm$0.12* & 0.567$\pm$0.19* & 0.618$\pm$0.16* \\
        & SurvPATH & 0.538$\pm$0.06* & 0.469$\pm$0.14* & 0.638$\pm$0.21* & 0.534$\pm$0.08*  \\
        & \modelacro \wo{}\(\mathcal{L}_{path}\) & 0.586$\pm$0.09* & 0.512$\pm$0.16* & 0.690$\pm$0.30* & 0.652$\pm$0.11* \\
        & \modelacro \wo{}$\mathcal{L}_{\text{sp}}$ & 0.620$\pm$0.10 & 0.743$\pm$0.27* & 0.743$\pm$0.27 & 0.662$\pm$0.12 \\
        & \modelacroNSP & \textbf{0.644$\pm$0.08} & 0.562$\pm$0.13* & \textbf{0.762$\pm$0.19} & \textbf{0.680$\pm$0.11}  \\
        \midrule
        \multirow{6}{*}{\rotatebox[origin=c]{90}{ADNI}} 
        & CCA & 0.592$\pm$0.04* & 0.242$\pm$0.11* & 0.810$\pm$0.05* & 0.526$\pm$0.05*  \\
        & G-MIND & 0.800$\pm$0.06* &  0.838$\pm$0.09* &  0.776$\pm$0.09* & 0.897$\pm$0.05* \\
        & SurvPATH & 0.818$\pm$0.06*  & \textbf{0.867$\pm$0.09}  & 0.787$\pm$0.08*   & \textbf{0.908$\pm$0.04}  \\
        & \modelacro \wo{}\(\mathcal{L}_{path}\) & 0.746$\pm$0.06* & 0.781$\pm$0.09* & 0.724$\pm$0.07* & 0.856$\pm$0.05* \\
        & \modelacro \wo{}$\mathcal{L}_{\text{sp}}$ & 0.758$\pm$0.05* &  0.800$\pm$0.09* & 0.731$\pm$0.06* & 0.858$\pm$0.05* \\
        & \modelacroNSP & \textbf{0.827$\pm$0.06} & 0.733$\pm$0.08* & \textbf{0.884$\pm$0.06} & 0.880$\pm$0.06* \\
        \bottomrule
    \end{tabular}}
\end{table}

\textbf{Quantitative Disease Prediction:} Table~\ref{tab:quantitative_results} reports the patient vs. control classification performance of the testing folds within a 10-fold cross-validation setup. As seen, \modelacro attains the highest accuracy and specificity for both the ACE and ADNI datasets. \modelacro also achieves the highest AUC on the ACE dataset and comparable performance to other deep learning models on the ADNI dataset. We note a drop in sensitivity for \modelacroNSP, which may be attributed to the small sizes of the encoding and classifier networks. We note that removing either the pathway similarity or sparsity loss in the ablation models diminishes performance, reinforcing the critical role of these terms in capturing robust interactions. While SurvPATH and G-MIND achieve competitive sensitivity and AUC on ADNI, they underperform in three of the four metrics on ACE, thus underscoring the generalizability of \modelacroNSP.

\begin{figure}[t]
    \includegraphics[width=\textwidth]{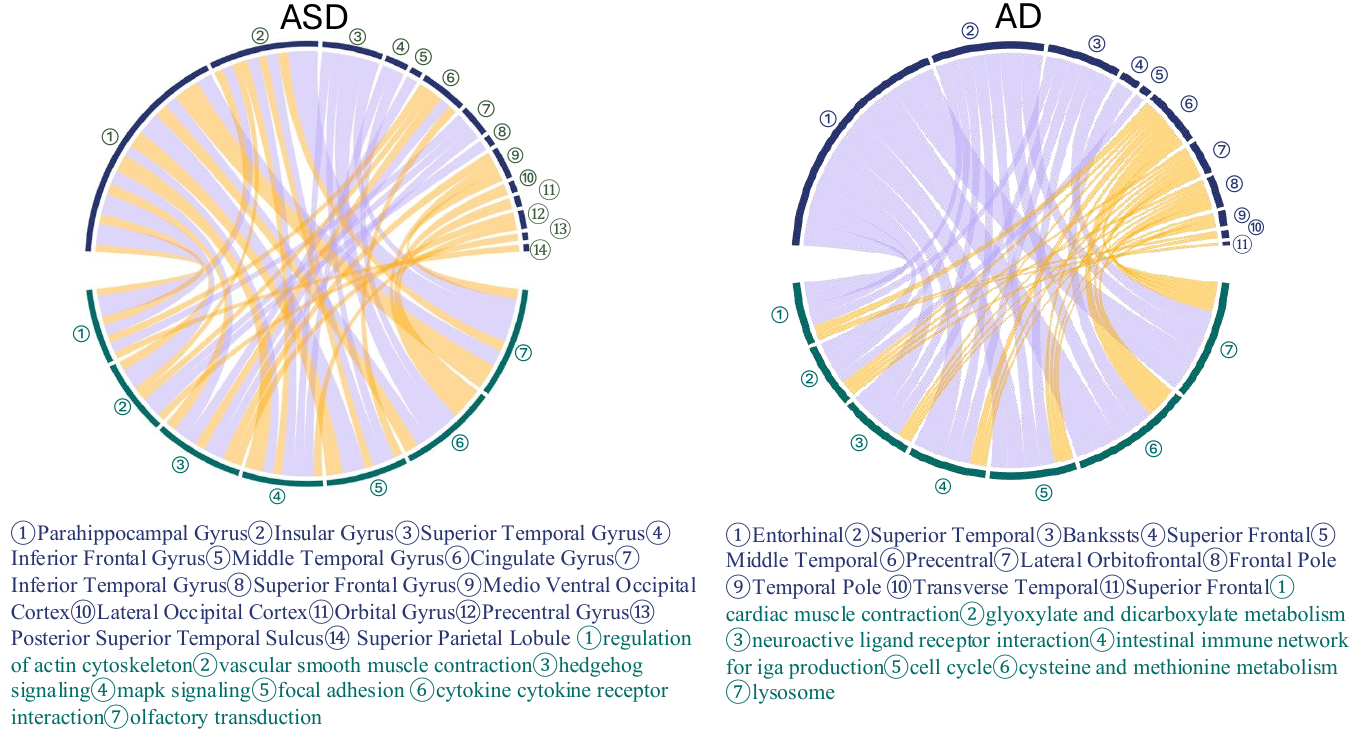}
    \caption{Visualization of the top brain-pathway associations learned by the mean attention matrices $\mathbf{\bar{A}}_{\text{PAT}}$ and $\bar{\mathbf{A}}_{\text{ND}}$ for ACE (left) and ADNI (right). The upper portion of each circle corresponds to brain ROIs, and the top portion corresponds to biological pathways. Purple connections are from $\mathbf{\bar{A}}_{\text{PAT}}$ and yellow connections are from $\bar{\mathbf{A}}_{\text{ND}}$.} 
    \label{fig:Result}
\end{figure}


\medskip \noindent
\textbf{Learned Imaging-Genetics Associations:}
To assess the interpretability of our model, we compute the mean attention matrices $\bar{\mathbf{A}}_{\text{PAT}}$ and $\bar{\mathbf{A}}_{\text{ND}}$ for the PAT and NC groups, respectively, across all testing folds in our 10-fold cross-validation. We then sum these matrices across columns to obtain pathway-level influences~$\mathbf{\bar{s}}_{\text{PAT}}$, $\mathbf{\bar{s}}_{\text{ND}}$ for both the ACE and ADNI datasets. We extract the intersection of the top seven pathways for each group (PAT and NC) for interpretation. Correspondingly, we extract four imaging ROIs that exhibit the highest values in the mean attention matrix for both PAT and NC groups.

Fig~\ref{fig:Result} right illustrates the predominant pathway–imaging feature interactions for the ACE and ADNI datasets. We observe that each pathway in the ACE dataset interacts with diverse brain ROIs, whereas in ADNI, each pathway is linked to more consistent ROIs. The patient and control groups in the ACE dataset also tend to select overlapping brain regions, suggesting a broader variety of neural alterations in autism. In contrast, for ADNI, the ROIs chosen by patients and controls diverge more noticeably, implying a greater separability between cohorts. These observations align with established clinical findings.

For autism (ACE), \modelacro highlights pathways involved in perception, neuro-inflammation, and cellular communication. These pathways interact with brain regions crucial for sensory processing, perception, emotion, and cognition. Such findings have been previously implicated in the ASD literature and support the biological relevance of our learned associations~\cite{hartig2021genetic,xu2015inflammatory}.
For Alzheimer's disease (ADNI), our model identifies pathways related to neuroinflammation and neurodegeneration, interacting with brain regions responsible for memory, language, and social functions. These associations are consistent with well-documented patterns of cortical atrophy and functional decline in AD~\cite{desikan2006automated}.



\section{Conclusion}  
We have presented \modelacroNSP, an explainable deep learning framework that integrates genetics data with MRI scans to elucidate the complex interactions between biological pathways and brain structure in neurological disorders. Our approach leverages an early fusion strategy with a cross-attention layer to effectively combine multimodal information. To enhance interpretability, we introduced two specialized loss functions: a sparsity loss to focus on the most critical interactions within the attention matrix, and (2) a pathway similarity loss to ensure consistent pathway identification across cohorts. We validated \modelacro on datasets for ASD and AD. Our results demonstrate that the proposed framework not only achieves high prediction accuracy but also uncovers biologically meaningful interactions between genetic pathways and brain regions.

\begin{credits}
\subsubsection{\ackname} This work was supported by the National Institutes of Health awards 1R01HD108790 (PI Venkataraman) and 1R01EB029977 (PI Caffo) and the National Science Foundation CAREER award 1845430 (PI Venkataraman).

\subsubsection{\discintname}
All authors disclosed no relevant relationships.
\end{credits}
%
%
%
\bibliographystyle{splncs04}
\bibliography{mybibliography}
%




\end{document}